\documentclass[letterpaper,journal]{IEEEtran}
\usepackage[T1]{fontenc}
\usepackage{amsmath,amssymb}
\usepackage{array}
\usepackage{booktabs}
\usepackage{graphicx}
\usepackage{stfloats} 
\usepackage{xcolor}
\usepackage{cite}
\usepackage[hyphens,spaces,obeyspaces]{url}
\usepackage{capt-of}
\usepackage[hidelinks]{hyperref}

\makeatletter
\def\url@codesty{\def\UrlFont{\ttfamily\small}}
\def\url@tabsty{\def\UrlFont{\ttfamily}}
\makeatother
\DeclareUrlCommand\code{\urlstyle{codesty}}   
\DeclareUrlCommand\tcode{\urlstyle{tabsty}}   
\newcommand{\func}[1]{\textit{\texttt{#1}}}   

\newcommand{\figbox}[4]{%
  \IfFileExists{#1}{\includegraphics[width=#2,height=#3,keepaspectratio]{#1}}{%
  {\setlength{\fboxsep}{0pt}\setlength{\fboxrule}{0.4pt}%
   \fcolorbox{black!35}{black!4}{\parbox[c][\dimexpr#3-0.8pt\relax][c]{\dimexpr#2-0.8pt\relax}{%
   \centering\scriptsize\color{black!60}\textbf{\detokenize{#1}}\\[2pt]%
   \parbox{0.86\linewidth}{\centering #4}}}}}}
\makeatletter
\newcounter{widefigure}
\let\c@widefigure\c@figure

\def\fps@widefigure{tbp}
\def\ftype@widefigure{4}
\def\ext@widefigure{lof}
\def\fnum@widefigure{\fnum@figure}
\newenvironment{widefigure}{\@dblfloat{widefigure}}{\end@dblfloat}
\makeatother

\begin{document}

\title{ExoStack: A Modular and Reconfigurable Software\\ Framework for Wearable Robotics Using ROS\,2}

\author{Debadrata~Sarkar, Upamanyu~Ray, Soumen~Sen, and~Aman~Arora%
\thanks{This work was supported by the CSIR-National Laboratory Scheme under ULIP [MLP002613] and the CSIR Mission Mode Project [NCP010303].}%
\thanks{This work involved human subjects. Approval of all ethical and experimental procedures was granted by the 7th sitting of the Institutional Ethical Board of CSIR-CMERI, Durgapur, India, on 27 February 2024.}%
\thanks{The authors are with the Human Centred Robotics and Cybernetics Group, CSIR--Central Mechanical Engineering Research Institute (CSIR-CMERI), Durgapur 713209, India. Corresponding author: Aman Arora (e-mail: aman.cmeri@nic.in).}}


\makeatletter
\def\@IEEEpubidpullup{1.3\baselineskip}
\makeatother
\IEEEpubid{\parbox{0.96\textwidth}{\centering\footnotesize\bfseries\color{red}%
This work has been submitted to the IEEE for possible publication.\\
Copyright may be transferred without notice, after which this version may no longer be accessible.}}

\maketitle

\nointerlineskip
\hbox to 0pt{\raisebox{2.0\baselineskip}[0pt][0pt]{%
\parbox{\textwidth}{\centering\large\bfseries\color{red}%
Code repository: \href{https://github.com/debadrata/ExoStack}{github.com/debadrata/ExoStack}\\[-1pt]
Documentation website: \href{https://debadrata.github.io/ExoStack_documentation/}{\nolinkurl{debadrata.github.io/ExoStack_documentation}}}}\hss}

\begin{abstract}
Wearable robots depend on a software chain that links body-worn sensing to gait interpretation, assistance generation, and actuation. Personalization further requires each measured response to remain associated with the assistance parameters that generated it. This paper presents ExoStack, a modular ROS\,2 framework for implementing that chain across embedded hubs and a host computer. Its Embedded Sensing and Actuation (ESA) library configures ESP32 boards as micro-ROS sensing and actuation hubs. ExoCore supplies multi-hub synchronization, gait-event detection, profile generation, motion planning, cost evaluation, and human-in-the-loop (HIL) optimization as configuration-instantiated nodes. The PASE desktop application edits configurations, launches the processing stages, and visualizes ROS\,2 topics. Hardware drivers and algorithm plug-ins can be added through defined interfaces without modifying the generic processing nodes. We characterize hub and planner timing on three embedded computers and demonstrate reuse on two pneumatic exosuits.
\end{abstract}


\section{Introduction}
\IEEEPARstart{W}{earable} assistive robots are progressing from laboratory demonstrations toward daily and clinical use~\cite{siviy2023,xiloyannis2022}. A central challenge is to deliver assistance that responds to the wearer rather than to a fixed gait template~\cite{molinaro2024,medrano2023}. Human-in-the-loop (HIL) optimization provides a systematic route to personalization: an optimizer proposes assistance parameters, the robot applies them during walking, and a cost measured on the wearer informs the next proposal~\cite{poggensee2021,slade2022,zhang2017,ding2018,diaz2023,slade2024}.

Implementing this loop requires coordinated functions beyond the joint controller itself. Sensor streams from multiple body-worn boards must be acquired and aligned. Gait events must trigger the appropriate assistance profile, and the measured cost must be assigned to the candidate and gait cycles under evaluation. In many research systems, these functions evolve as part of a device-specific application. A change in sensor layout, assisted joint, actuator, or optimization strategy can then propagate through acquisition, control, and visualization code, making reuse and reproducibility difficult.

Existing platforms address important parts of wearable-robot development. OpenExo~\cite{williams2025} provides a modular exoskeleton hardware and control platform, whereas Epically Powerful~\cite{leestma2025} offers a Python infrastructure for quasi-direct-drive actuators, sensing, and visualization. Their design choices reflect their target hardware and control workflows. ExoStack addresses a complementary need: a configurable pipeline for systems in which sensors and actuators are distributed across several body-worn microcontrollers and host-side algorithms may change between experiments. ROS\,2~\cite{macenski2022} supplies the host-side communication model, while micro-ROS~\cite{belsare2023} extends ROS\,2 communication to resource-constrained boards. Mature ROS\,2 frameworks such as Nav2~\cite{macenski2020}, MoveIt~\cite{coleman2014}, and ros\_control~\cite{chitta2017} demonstrate the value of separating reusable infrastructure from robot-specific configuration. ExoStack applies this principle to the sensing-to-assistance workflow of wearable robotics.
\IEEEpubidadjcol

To the best of our knowledge, ExoStack is the first ROS~2 framework to cover the complete wearable-robot pipeline, from body-worn sensing to personalized assistance. Its contributions are: (1) a firmware API for distributed, multi-hub sensing and actuation, publishing analog, digital, and I\textsuperscript{2}C readings on a shared time base without hand-written micro-ROS code; (2) configurable ROS~2 nodes for gait-event detection, motion planning, and assistance deployment or optimization, instantiated per limb, joint, and event; (3) a coordinated HIL loop linking each optimizer candidate to its execution and cost measured over its gait cycles, preventing cost misattribution; (4) a unified profile framework routing speed- and intent-indexed lookup tables and HIL candidates through the same planners; (5) template-guided extensibility through self-registering drivers and plug-in algorithms; and (6) a graphical tool to configure, launch, and visualize the stack. Timing benchmarks and two pneumatic-exosuit implementations validate the framework.

\setcounter{figure}{1}
\begin{widefigure}[!b]
\centering
\includegraphics[scale=0.1]{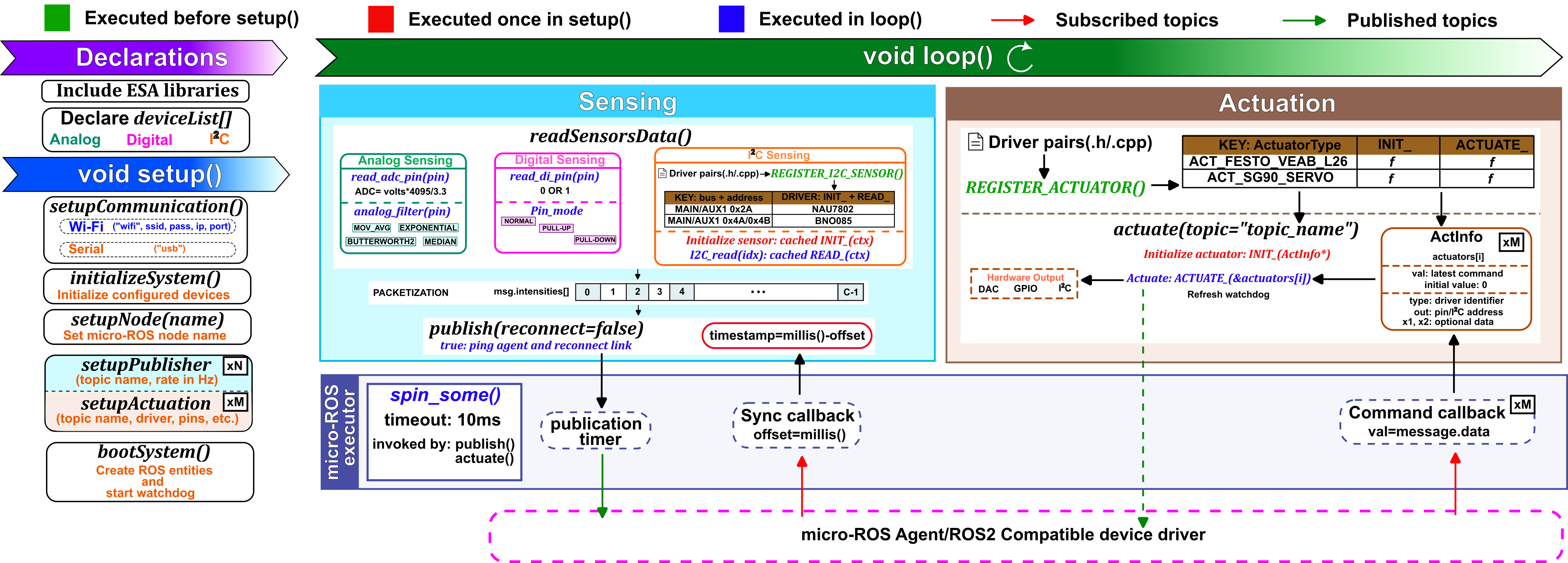}
\caption{ESA hub sequence. Drivers register before setup. Setup initializes communication, devices, the ROS node, publishers, and actuators. The runtime loop publishes sensor channels in \texttt{msg.intensities[]} and applies actuator commands while the executor services timers and callbacks.}
\label{fig:esa}
\end{widefigure}

\section{System Overview}
ExoStack divides the system into on-body hubs, host-side processing, and a user-facing configuration tool (Fig.~\ref{fig:overview}). ESA runs on ESP32 boards and exposes each sensing or actuation hub to ROS\,2 through a micro-ROS agent using Wi-Fi or USB. On the host, ExoCore consumes synchronized sensor channels, detects gait events, constructs assistance profiles, and publishes actuator set-points. When HIL is enabled, cost evaluators and optimizers form a feedback path around the same profile and planner nodes. Four YAML files define the hubs and channel order, event detectors, planners, and profile sources. Launch files create node instances from these entries. Device drivers are resolved through the ESA registries, and ExoCore algorithms are loaded from plug-in directories. PASE edits the YAML files, starts the agents and processing stages, and visualizes topic data. External ROS\,2 estimators and actuator drivers can connect through the published topic interfaces.

\setcounter{figure}{0}
\begin{figure}[!t]
\centering
\figbox{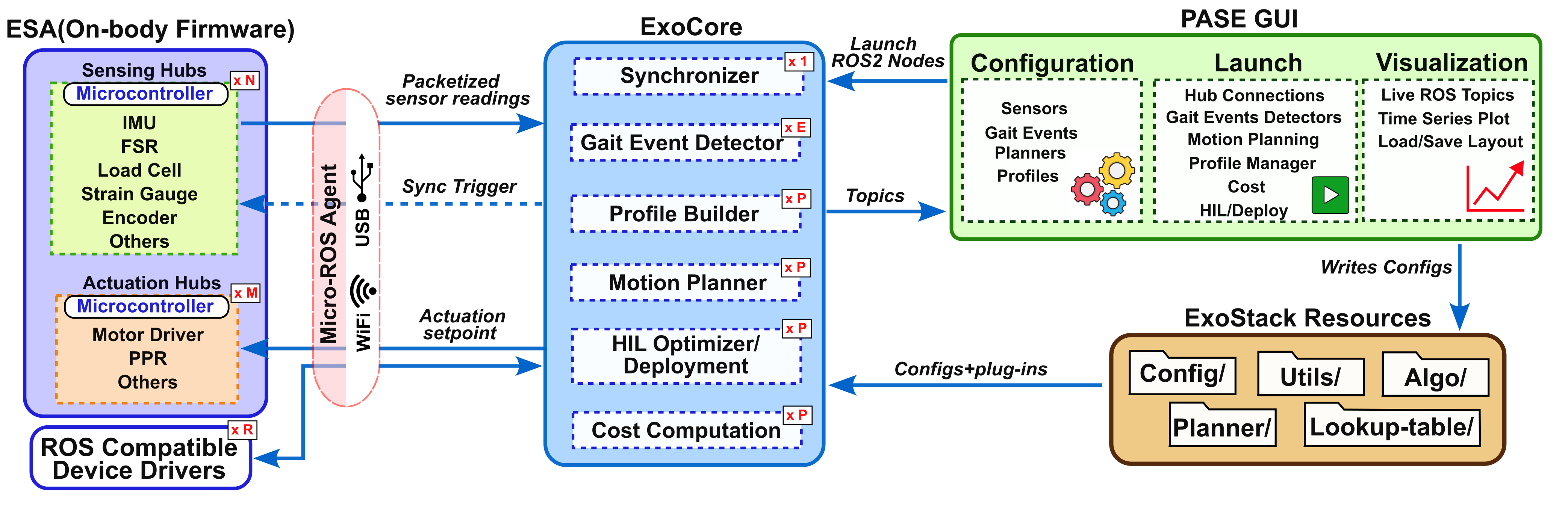}{\columnwidth}{1.12in}{System overview.}
\caption{System overview. ESA hubs exchange sensor data and set-points with ExoCore through micro-ROS. PASE configures and launches the graph and visualizes topics. YAML entries determine the hub, driver, detector, planner, profile, cost, and HIL instances.}
\label{fig:overview}
\end{figure}
\setcounter{figure}{2}

\begin{figure*}[!t]
\centering
\includegraphics[scale=0.1]{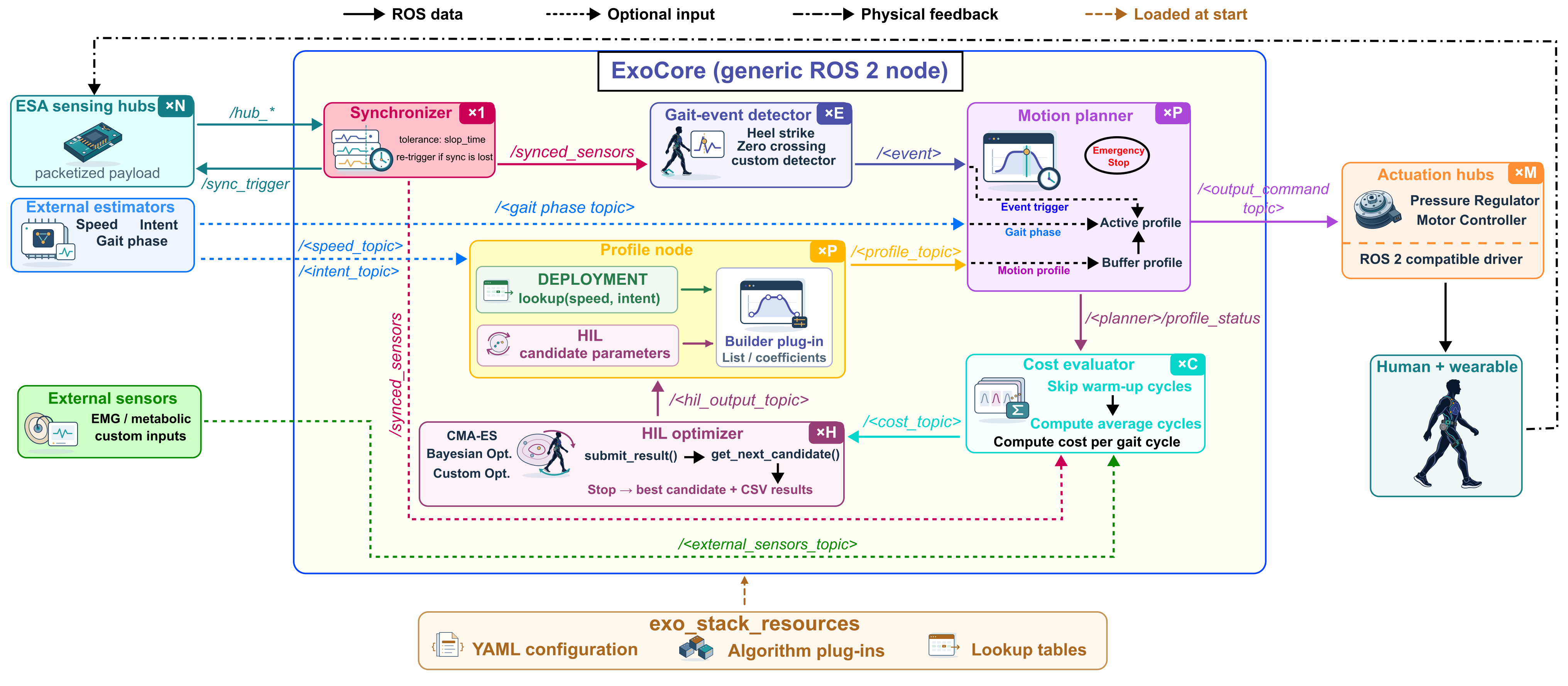}
\caption{ExoCore, the generic ROS\,2 nodes, and their topics. $\times$E, $\times$P, $\times$C, and $\times$H count the instances created from YAML entries. Solid arrows carry ROS data, dotted arrows optional inputs, the dash-dotted loop the physical feedback through the wearer, and dashed arrows what is loaded at start. The candidate number $k$ travels in the profile, status, and cost messages, so the optimizer accepts only the cost of the candidate it proposed.}
\label{fig:exocore}
\end{figure*}

\begin{figure}[!b]
\centering
\figbox{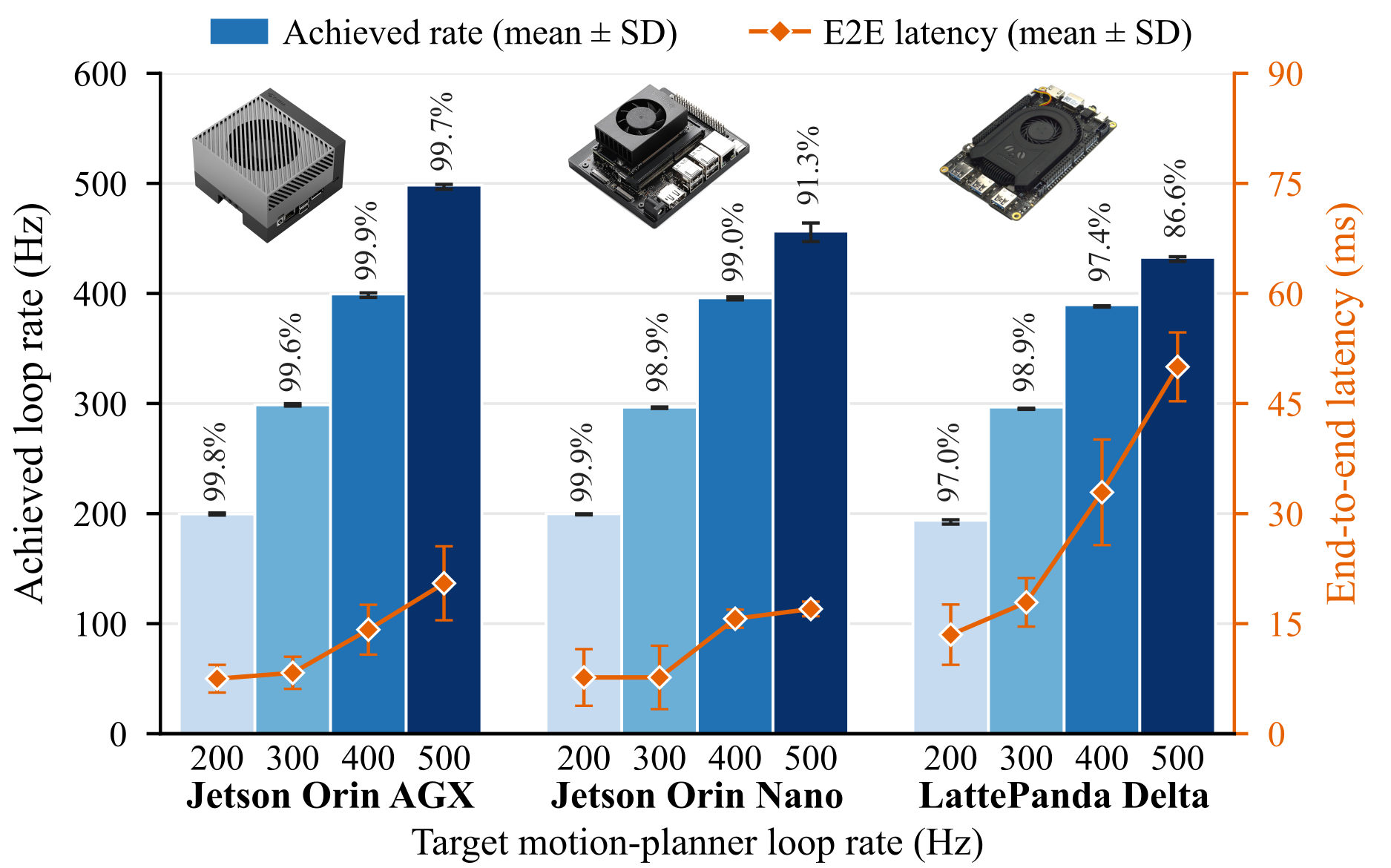}{0.94\columnwidth}{1.55in}{Benchmark results.}
\caption{Planner target-rate attainment and end-to-end latency on the three host computers. Values are mean $\pm$ SD.}
\label{fig:bench}
\end{figure}

\begin{figure*}[!t]
\centering
\figbox{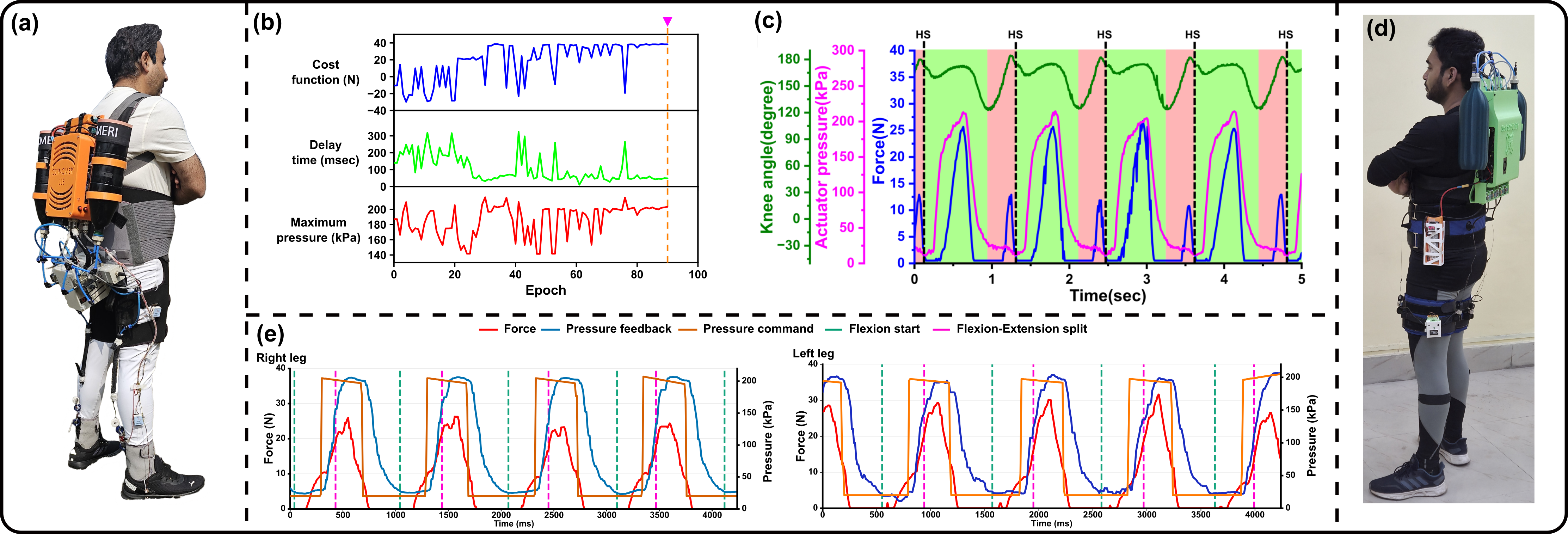}{0.74\textwidth}{1.92in}{Validation on two exosuits.}
\caption{Validation on (a) the bilateral knee-flexion exosuit SoftEXOwalk, (b) its Bayesian HIL optimization, and (c) the resulting profiles and transmitted forces, and (d) the hip flexion/extension exosuit with (e) example CMA-ES-tuned assistance profiles.}
\label{fig:deploy}
\end{figure*}

\section{Embedded Sensing and Actuation (ESA)}
\subsection{Declaring a Hub}

ESA is an Arduino library that turns an ESP32 board into a sensing hub, an actuation hub, or both. The sketch declares sensors in \code{deviceList[]} using an interface type (\code{AI}, \code{DI}, or \code{I2C}), a GPIO pin or bus address, a channel label, and type-specific settings. These settings select analog filtering, digital input mode, or the I\textsuperscript{2}C bus and device parameters (Sup.~Table~S1). The initialization sequence is explicit. \func{setupCommunication()} selects Wi-Fi or USB, \func{initializeSystem()} initializes the declared devices, and \func{setupNode()} assigns the ROS node name. \func{setupPublisher()} and \func{setupActuation()} declare the data and command interfaces. Finally, \func{bootSystem()} creates the micro-ROS entities and starts the 3-s watchdog (Fig.~\ref{fig:esa}).

\subsection{Sensing and Actuation}

At run time, \func{readSensorsData()} reads the declared inputs. Analog channels support moving-average, median, exponential, and second-order Butterworth low-pass, high-pass, band-pass, and notch filters. Digital channels support normal, pull-down, and pull-up modes. I\textsuperscript{2}C acquisition is delegated to the registered device drivers. Values are serialized in declaration order into \code{sensor_msgs/LaserScan.intensities[]} and published by a micro-ROS timer at the configured rate. Reusing a standard message type avoids maintaining a custom micro-ROS interface. An actuation hub subscribes to one \code{Float32} command topic per actuator and passes the latest set-point to the corresponding driver through \func{actuate()}. The sensing and actuation loops service the micro-ROS executor with a bounded 10-ms spin, allowing timers and command callbacks to be handled alongside the application loop.

When reconnection monitoring is requested, \func{publish(true)} also pings the agent and recreates the ROS entities after a lost connection. The watchdog resets a stalled application loop, while a status LED indicates ROS entity-creation failures. Filter type and parameters are selected per analog channel, and the filter coefficients are initialized using the publication rate. This keeps device-specific acquisition settings in the sketch rather than in the host-side processing code.

\subsection{Extending the Driver Registry}

Hardware-specific code is kept in paired header and source driver files. An I\textsuperscript{2}C sensor driver uses \mbox{\code{REGISTER_I2C_SENSOR}} to register initialization and read callbacks. ESA resolves the callback by bus and address during initialization. Actuator drivers use \mbox{\code{REGISTER_ACTUATOR}} to register initialization and command callbacks for an \code{ActuatorType}. The sketch then binds an actuator instance to a command topic and its hardware pins or address using \func{setupActuation(topic, type, output, extra1, extra2)}. The supplied interfaces cover the BNO085 IMU, NAU7802 load-cell amplifier, AS5600 magnetic encoder, Festo VEAB-L-26 pressure regulator, and SG90 servo. New hardware can be added through the same registries, or exposed through an independent ROS\,2 driver when embedded integration is unnecessary.

\subsection{Synchronizing the Hubs}

The synchronizer periodically broadcasts \code{/sync_trigger} while establishing a set of active hubs. Each hub records the local \func{millis()} value when the trigger arrives and stamps subsequent messages with elapsed time from that event. These trigger-relative stamps place the hub streams on a comparable time base without synchronizing their local clocks. The host matches incoming messages within a configurable timestamp tolerance and monitors the stream for loss of synchronization (Sec.~\ref{sec:sync}).

\section{ExoCore}
ExoCore is the host-side ROS\,2 Python package (Fig.~\ref{fig:exocore}). The implementation targets Ubuntu 22.04 and ROS\,2 Humble, with Python 3.10 on the host and the Humble micro-ROS client on ESP32 boards. It comprises six executables: \code{sync}, \code{gait_event_detector}, \code{mplanner}, \code{actuation_profile}, \code{cost}, and \code{hil}. Launch files instantiate detector, planner, profile, cost, and optimizer nodes from entries in \code{sensors.yaml}, \code{gait_events.yaml}, \code{mplanner.yaml}, and \code{profile.yaml}. The configuration therefore determines both the number of node instances and the signals and topics that connect them. For example, separate flexion- and extension-onset events for each hip create four detector instances, while each assisted motion receives its own planner and profile node. Cost and HIL nodes are created only for profiles that enable those functions. The ROS\,2 topic names, message types, and array layouts are documented in Sup.~Table~S3.

\subsection{Synchronization and Gait Events}\label{sec:sync}

The synchronizer publishes \code{/sync_trigger} at 5~Hz while it is in trigger mode. It subscribes to the configured hub topics using an approximate-time synchronizer with a user-defined \code{slop_time}. Once a matched set is received, it concatenates the \code{LaserScan.intensities[]} arrays in the hub and channel order specified by \code{sensors.yaml} and publishes the result as \code{Float32MultiArray} on \code{/synced_sensors}. If no matched set arrives for 1~s, the node returns to trigger mode and resets its synchronizer. The associated \code{SensorUtil} maps each hub--channel pair to its position in the concatenated array, so algorithms refer to named signals rather than hard-coded indices.

Each gait-event node selects its configured channels from \code{/synced_sensors}, passes them to a dynamically loaded \func{detect()} plug-in, and publishes an \code{Int32} event when the detector fires. The supplied plug-ins include a thresholded rising-edge detector for foot-pressure signals (\code{HeelStrike}) and positive-to-negative or negative-to-positive zero-crossing detectors for angular signals. Each detector maintains a sliding window of $N$ samples and requires a configurable fraction $F$ of consecutive samples at the end of the window to support the new state. After an event, it rearms only when the window has returned fully to the opposite state (Sup.~Fig.~S1a).

\subsection{Motion Planning and Assistance-Profile Generation}

Each assisted motion uses a profile node and planner to convert gait events and wearer state into actuator set-points. In \code{Deployment}, a JSON lookup table supplies parameters indexed by walking speed and, optionally, locomotion intent. A lookup plug-in selects the nearest speed entry or interpolates between entries, and a builder converts the parameters into time--value waypoints or function coefficients. In \code{HIL}, the same builder encodes optimizer candidates. Profiles are published as \code{Float32MultiArray} messages with a common layout comprising candidate identifier $k$ (zero in Deployment), type, start, payload, end, and stop flag. The type distinguishes waypoint and coefficient representations, while the payload contains the corresponding values (Sup.~Fig.~S1b; Sup.~Sec.~2.2).

The planner evaluates the profile at its configured rate using time since the triggering gait event (\code{time_based}) or externally estimated gait phase (\code{gait_phase_based}). Incoming profiles are buffered until the next gait event. Outside the profile interval, or when an intent value of zero raises the emergency-stop flag, it publishes \code{default_output}. Manual mode adds an offset to \code{default_output}. The supplied working planner uses linear interpolation; PCHIP is a separate, smoother, non-overshooting alternative, but is documented as non-operational in v1.0~\cite{fritsch1980} (Sup.~Fig.~S1c).

\subsection{Human-in-the-Loop Optimization}

HIL optimization evaluates one candidate at a time (Fig.~\ref{fig:exocore}). The optimizer publishes a parameter vector with identifier $k$. The profile node builds it, and the planner activates it at the next gait event. At each subsequent event, the planner reports $k$ and the cycle count. The cost node discards the configured adaptation cycles, evaluates its plug-in over the next specified cycles, and returns the mean cost with the same identifier. The optimizer accepts only the outstanding candidate's result, preventing delayed feedback from being assigned to a later proposal. Cost plug-ins can use synchronized channels and external signals such as metabolic or EMG measurements. ExoStack includes CMA-ES (\code{OPT_CMA_ES})~\cite{hansen2016,nomura2024} and Bayesian optimization (\code{OPT_BAYESIAN})~\cite{nogueira2014}. Candidate parameters and costs are saved as CSV.

\section{PASE: Configure, Launch, and Visualize}

PASE is a PyQt5 desktop application with three linked functions. The Configuration tab edits the four YAML files and populates algorithm selectors from the available plug-in directories. The Launch tab starts the micro-ROS agents and ExoCore stages in dependency order and reports which configured nodes are active. The Visualization tab provides live plots of selected ROS\,2 topic fields, allows channels to be selected by their configured names, and overlays gait events on sensor traces. Plot arrangements can be saved and restored, supporting repeatable inspection during setup and experiments.

\section{Validation and Benchmarking}

\emph{Validation on two exosuits:} ExoStack was evaluated on two CSIR-CMERI pneumatic exosuits (Fig.~\ref{fig:deploy}, Sup.~Sec.~4), both using silicone-tube artificial muscles (pSTAMs), proportional pressure regulators, and the same ESA library and ExoCore executables. Only device lists, YAML configurations, and plug-ins differed. SoftEXOwalk~\cite{sarkar2026} provides bilateral knee-flexion assistance. Each leg used a USB sensing hub with an insole FSR, two BNO085 IMUs, a load cell, and a pSTAM strain sensor, and an actuation hub with a regulator and pressure feedback. FSR heel strikes triggered time-based planners, while Bayesian HIL tuned the amplitude and onset delay of a trapezoidal profile. Its cost plug-in estimated knee angle as $180^\circ+\theta_{\mathrm{thigh}}-\theta_{\mathrm{shank}}$, segmented force data at minimum knee angle, rewarded peak assistive force, and penalized peak resistive force above 15~N.

The second system assisted bilateral hip flexion and extension using four backpack-mounted pSTAMs coupled to the thighs through tendons and a speed-enhancement mechanism. Two wireless thigh hubs acquired insole-pressure, IMU, and load-cell signals. Two USB backpack hubs acquired encoder, muscle-strain, and regulator-pressure signals, while two USB actuation hubs controlled the regulators and exhaust valves. Opposite-polarity zero crossings of thigh angular velocity detected terminal flexion and extension. Four time-based planners generated bilateral commands, with CMA-ES optimizing onset delay, rise, hold, and fall times to maximize the peak assistive-to-resistive force ratio, subject to a 15~N peak-resistive-force limit. Both systems operated sensing and actuation at 100~Hz. Deployment profiles were indexed by leg, assisted motion, and walking speed from 3.5 to 5.5~km/h in 0.5-km/h increments.

\emph{Benchmarking:} Tests on ROS~2 Humble using a Jetson AGX Orin, Jetson Orin Nano, and LattePanda Delta confirmed stable distributed sensing, with eight hubs publishing at 100~Hz at a measured rate of 99.5$\pm$0.018~Hz. End-to-end tests further showed that all three hosts reliably support 200--300~Hz planning, maintaining at least 97\% of the target rate with latency below 18~ms. Performance becomes hardware-dependent at higher rates (Fig.~\ref{fig:bench}). The full benchmark protocol is provided in Sup.~Sec.~3.

\emph{Conclusion:} ExoStack links distributed sensing, gait-event assistance, and HIL personalization through configurable ROS\,2 nodes. Reuse across two pneumatic exosuits and timing tests on three embedded hosts demonstrate its practical value.

\clearpage
\twocolumn[{%
\section*{Description of Supplementary Material}
\vspace{2pt}
\centering
\figbox{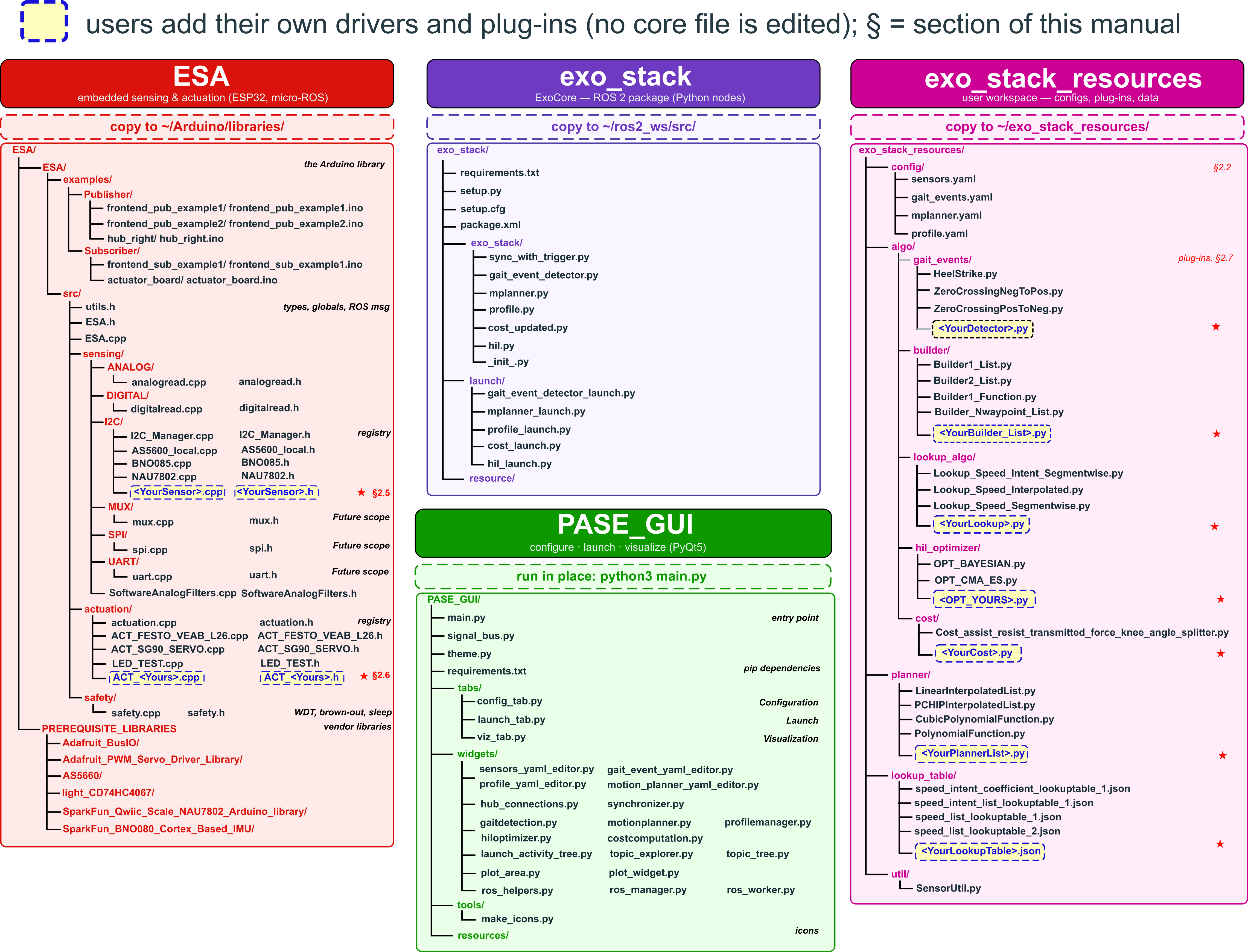}{\textwidth}{4.6in}{Repository layout (FIG8).}
\captionof{figure}{Repository layout of ExoStack. Each module is copied to the folder shown under its name. Dashed entries mark locations where users add their own drivers and plug-ins, with labels pointing to the relevant section of the supplementary document.}\label{fig:repo}
\vspace{10pt}}]

The supplementary material provides the source code and a companion document that serves as the user manual (Table~\ref{tab:supp}). The documentation website, \url{https://debadrata.github.io/ExoStack_documentation/}, provides configuration references, PASE screenshots, and videos of both exosuits. The code and documentation are available from the public repository \url{https://github.com/debadrata/ExoStack}. Its directory structure is shown in Fig.~\ref{fig:repo}. The software is distributed under the PolyForm Noncommercial License 1.0.0, and the manual under CC BY-NC 4.0.

Each module is installed where it runs. ESA is an Arduino library on the computer that programs the hubs, ExoCore is the ROS\,2 package \code{exo_stack} in a colcon workspace on the host computer, \code{exo_stack_resources} resides in the user's home directory, and PASE is a stand-alone Python application. The supplementary document follows the order in which a new user brings up a device: installation, hub firmware, node configuration, operation through PASE, and extension through templates. It also documents the benchmarking set-up and results (\S3) and the hardware, sensing, actuation, and control configurations used for both exosuit validations in Sec.~VI.

\begin{table}[!h]
\caption{Contents of the Supplementary Material}
\label{tab:supp}
\centering\footnotesize
\setlength{\tabcolsep}{3pt}
\begin{tabular}{@{}>{\raggedright\arraybackslash}p{1.02in}>{\raggedright\arraybackslash}p{2.38in}@{}}
\toprule
\textbf{Item} & \textbf{Content} \\
\midrule
\multicolumn{2}{@{}l}{\textit{Code} (PolyForm Noncommercial License 1.0.0)}\\
\tcode{ESA/} & Arduino library for ESP32 hubs: API, drivers, driver templates, example sketches\\
\tcode{exo_stack/} & ExoCore ROS\,2 package: six nodes and their launch files\\
\tcode{exo_stack_resources/} & YAML files, plug-ins and plug-in templates, lookup tables, HIL logs\\
\tcode{PASE_GUI/} & configuration, launch, and visualization tool\\
\midrule
\multicolumn{2}{@{}l}{\textit{Supplementary document} (PDF, CC BY-NC 4.0)}\\
\S1 & repository layout (Fig.~\ref{fig:repo})\\
\S2.1 & installation of ESA, the micro-ROS agent, ExoCore, and PASE\\
\S2.2 & configuration: hub sketches with the device list (Table~S1) and driver registries (Table~S2), YAML files, lookup tables, detection and profile examples (Fig.~S1)\\
\S2.3--2.4 & running the stack, including PASE launch and visualization\\
\S2.5--2.7 & driver and plug-in templates\\
\S3 & benchmarking set-up and results (Fig.~S2)\\
\S4 & configurations of the two validated exosuits\\
App.~A & troubleshooting and ROS\,2 interfaces (Table~S3)\\
\bottomrule
\end{tabular}
\end{table}

\clearpage
\raggedbottom
\bibliographystyle{IEEEtran}
\bibliography{refs}

\begin{thebibliography}{10}
\providecommand{\url}[1]{#1}
\csname url@samestyle\endcsname
\providecommand{\newblock}{\relax}
\providecommand{\bibinfo}[2]{#2}
\providecommand{\BIBentrySTDinterwordspacing}{\spaceskip=0pt\relax}
\providecommand{\BIBentryALTinterwordstretchfactor}{4}
\providecommand{\BIBentryALTinterwordspacing}{\spaceskip=\fontdimen2\font plus
\BIBentryALTinterwordstretchfactor\fontdimen3\font minus
  \fontdimen4\font\relax}
\providecommand{\BIBforeignlanguage}[2]{{%
\expandafter\ifx\csname l@#1\endcsname\relax
\typeout{** WARNING: IEEEtran.bst: No hyphenation pattern has been}%
\typeout{** loaded for the language `#1'. Using the pattern for}%
\typeout{** the default language instead.}%
\else
\language=\csname l@#1\endcsname
\fi
#2}}
\providecommand{\BIBdecl}{\relax}
\BIBdecl

\bibitem{siviy2023}
C.~Siviy, L.~M. Baker, B.~T. Quinlivan, F.~Porciuncula, K.~Swaminathan, L.~N.
  Awad, and C.~J. Walsh, ``Opportunities and challenges in the development of
  exoskeletons for locomotor assistance,'' \emph{Nat. Biomed. Eng.}, vol.~7,
  no.~4, pp. 456--472, 2023.

\bibitem{xiloyannis2022}
M.~Xiloyannis, R.~Alicea, A.-M. Georgarakis, F.~L. Haufe, P.~Wolf, L.~Masia,
  and R.~Riener, ``Soft robotic suits: State of the art, core technologies, and
  open challenges,'' \emph{IEEE Trans. Robot.}, vol.~38, no.~3, pp. 1343--1362,
  2022.

\bibitem{molinaro2024}
D.~D. Molinaro, K.~L. Scherpereel, E.~B. Schonhaut, G.~Evangelopoulos, M.~K.
  Shepherd, and A.~J. Young, ``Task-agnostic exoskeleton control via biological
  joint moment estimation,'' \emph{Nature}, vol. 635, pp. 337--344, 2024.

\bibitem{medrano2023}
R.~L. Medrano, G.~C. Thomas, C.~G. Keais, E.~J. Rouse, and R.~D. Gregg,
  ``Real-time gait phase and task estimation for controlling a powered ankle
  exoskeleton on extremely uneven terrain,'' \emph{IEEE Trans. Robot.},
  vol.~39, no.~3, pp. 2170--2182, 2023.

\bibitem{poggensee2021}
K.~L. Poggensee and S.~H. Collins, ``How adaptation, training, and
  customization contribute to benefits from exoskeleton assistance,''
  \emph{Sci. Robot.}, vol.~6, no.~58, p. eabf1078, 2021.

\bibitem{slade2022}
P.~Slade, M.~J. Kochenderfer, S.~L. Delp, and S.~H. Collins, ``Personalizing
  exoskeleton assistance while walking in the real world,'' \emph{Nature}, vol.
  610, pp. 277--282, 2022.

\bibitem{zhang2017}
J.~Zhang, P.~Fiers, K.~A. Witte, R.~W. Jackson, K.~L. Poggensee, C.~G. Atkeson,
  and S.~H. Collins, ``Human-in-the-loop optimization of exoskeleton assistance
  during walking,'' \emph{Science}, vol. 356, no. 6344, pp. 1280--1284, 2017.

\bibitem{ding2018}
Y.~Ding, M.~Kim, S.~Kuindersma, and C.~J. Walsh, ``Human-in-the-loop
  optimization of hip assistance with a soft exosuit during walking,''
  \emph{Sci. Robot.}, vol.~3, no.~15, p. eaar5438, 2018.

\bibitem{diaz2023}
M.~A. D{\'i}az \emph{et~al.}, ``Human-in-the-loop optimization of wearable
  robotic devices to improve human--robot interaction: A systematic review,''
  \emph{IEEE Trans. Cybern.}, vol.~53, no.~12, pp. 7483--7496, 2023.

\bibitem{slade2024}
P.~Slade, C.~Atkeson, J.~M. Donelan, H.~Houdijk, K.~A. Ingraham, M.~Kim,
  K.~Kong, K.~L. Poggensee, R.~Riener, M.~Steinert, J.~Zhang, and S.~H.
  Collins, ``On human-in-the-loop optimization of human--robot interaction,''
  \emph{Nature}, vol. 633, pp. 779--788, 2024.

\bibitem{williams2025}
J.~R. Williams, C.~F. Cuddeback, S.~Fang, D.~Colley, N.~Enlow, P.~Cox,
  P.~Pridham, and Z.~F. Lerner, ``{OpenExo}: An open-source modular exoskeleton
  to augment human function,'' \emph{Sci. Robot.}, vol.~10, p. eadt1591, 2025.

\bibitem{leestma2025}
J.~K. Leestma, S.~R. Nathella, C.~P.~O. Nuesslein, S.~Mathur, G.~S. Sawicki,
  and A.~J. Young, ``Epically {Powerful}: An open-source software and
  mechatronics infrastructure for wearable robotic systems,'' arXiv:2511.05033,
  2025.

\bibitem{macenski2022}
S.~Macenski, T.~Foote, B.~Gerkey, C.~Lalancette, and W.~Woodall, ``Robot
  {Operating} {System} 2: Design, architecture, and uses in the wild,''
  \emph{Sci. Robot.}, vol.~7, no.~66, p. eabm6074, 2022.

\bibitem{macenski2020}
S.~Macenski, F.~Mart{\'i}n, R.~White, and J.~{Gin{\'e}s Clavero}, ``The
  {Marathon} 2: A navigation system,'' in \emph{Proc. IEEE/RSJ Int. Conf.
  Intell. Robots Syst. (IROS)}, 2020.

\bibitem{coleman2014}
D.~Coleman, I.~A. {\c{S}}ucan, S.~Chitta, and N.~Correll, ``Reducing the
  barrier to entry of complex robotic software: A {MoveIt!} case study,''
  \emph{J. Softw. Eng. Robot.}, vol.~5, no.~1, pp. 3--16, 2014.

\bibitem{chitta2017}
S.~Chitta \emph{et~al.}, ``ros\_control: A generic and simple control framework
  for {ROS},'' \emph{J. Open Source Softw.}, vol.~2, no.~20, p. 456, 2017.

\bibitem{belsare2023}
K.~Belsare \emph{et~al.}, ``Micro-{ROS},'' in \emph{Robot Operating System
  (ROS): The Complete Reference}, A.~Koubaa, Ed.\hskip 1em plus 0.5em minus
  0.4em\relax Cham, Switzerland: Springer, 2023, vol.~7, pp. 3--55.

\bibitem{sarkar2026}
D.~Sarkar, S.~Chakraborty, S.~Sen, S.~S. Roy, and A.~Arora, ``{SoftEXOwalk}: A
  human-in-the-loop optimized soft exosuit for assisting knee flexion using
  sensorized pneumatic muscles,'' \emph{IEEE Trans. Med. Robot. Bionics}, 2026.

\bibitem{fritsch1980}
F.~N. Fritsch and R.~E. Carlson, ``Monotone piecewise cubic interpolation,''
  \emph{SIAM J. Numer. Anal.}, vol.~17, no.~2, pp. 238--246, 1980.

\bibitem{hansen2016}
N.~Hansen, ``The {CMA} evolution strategy: A tutorial,'' arXiv:1604.00772,
  2016.

\bibitem{nomura2024}
M.~Nomura and M.~Shibata, ``cmaes: A simple yet practical {Python} library for
  {CMA-ES},'' arXiv:2402.01373, 2024.

\bibitem{nogueira2014}
F.~Nogueira, ``Bayesian {Optimization}: Open source constrained global
  optimization tool for {Python},'' [Online]. Available:
  \url{https://github.com/bayesian-optimization/BayesianOptimization}, 2014.

\end{thebibliography}

\end{document}